\documentclass[%
 aip,
 amsmath,
 amssymb,
reprint,%
floatfix,
]{revtex4-1}

\usepackage{graphicx}
\usepackage{dcolumn}
\usepackage{bm}

\usepackage{algpseudocode,algorithm,algorithmicx}
\usepackage{hyperref}

\usepackage[utf8]{inputenc}
\usepackage[T1]{fontenc}
\usepackage{mathptmx}
\usepackage{xcolor}

\begin{document}

\title[]{Machine learning enabling high-throughput and remote operations at large-scale user facilities}

\author{Tatiana~Konstantinova}
\thanks{These two authors contributed equally}
\affiliation{National Synchrotron Light Source II, Brookhaven National Laboratory, Upton, NY 11973, USA}

\author{Phillip M. Maffettone}
\thanks{These two authors contributed equally}
\affiliation{National Synchrotron Light Source II, Brookhaven National Laboratory, Upton, NY 11973, USA}

\author{Bruce~Ravel}
\affiliation{National Institute of Standards and Technology, Gaithersburg, MD 20899, USA}

\author{Stuart~I. Campbell}
\affiliation{National Synchrotron Light Source II, Brookhaven National Laboratory, Upton, NY 11973, USA}

\author{Andi~M.~Barbour}
\affiliation{National Synchrotron Light Source II, Brookhaven National Laboratory, Upton, NY 11973, USA}

\author{Daniel Olds}
 \email{dolds@bnl.gov}
\affiliation{National Synchrotron Light Source II, Brookhaven National Laboratory, Upton, NY 11973, USA}%

\date{\today}

\begin{abstract}
Imaging, scattering, and spectroscopy are fundamental in understanding and discovering new functional materials. 
Contemporary innovations in automation and experimental techniques have led to these measurements being performed much faster and with higher resolution, thus producing vast amounts of data for analysis. 
These innovations are particularly pronounced at user facilities and synchrotron light sources. 
Machine learning (ML) methods are regularly developed to process and interpret large datasets in real-time with measurements. 
However, there remain conceptual barriers to entry for the facility general user community, whom often lack expertise in ML, and technical barriers for deploying ML models.
Herein, we demonstrate a variety of archetypal ML models for on-the-fly analysis at multiple beamlines at the National Synchrotron Light Source II (NSLS-II). 
We describe these examples instructively, with a focus on integrating the models into existing experimental workflows, such that the reader can easily include their own ML techniques into experiments at NSLS-II or facilities with a common infrastructure. 
The framework presented here shows how with little effort, diverse ML models operate in conjunction with feedback loops \emph{via} integration into the existing Bluesky Suite for experimental orchestration and data management. 
\end{abstract}

\maketitle


\section{Introduction}

The past decade has seen a surge in the use of artificial intelligence (AI) and machine learning (ML) across the sciences.
These tools have become essential for interpreting increasingly large datasets, which are simply too massive to be effectively analyzed manually.
Not only has AI increased the ability to interpret these datasets, it has increased the pace at which decisions are made,\cite{Zhou_2017} and in some cases outperforms human expertise.\cite{Brown_2019}
Applications of AI have enabled significant strides in physics, \cite{Mehta_2019} chemistry, \cite{Gromski_2019, Butler_2018} materials science,\cite{Batra_2020} and biology.\cite{Senior_2020}
It is thus unsurprising that light sources and central facilities have begun to look toward these technologies for active decision making, experimental monitoring, and guided physics simulations.\cite{Campbell_2020, Duris_2020, Ren_2018}
In the following, we outline the challenges that are necessitating the adoption of AI, describe how to navigate the barrier to entry as a scientist, and demonstrate some archetypal uses of AI and exemplify their facile deployment at a variety of experiments across a central facility using the Bluesky software suite.\cite{Bluesky}
\par

The pressing need to create new tools to optimize human effort at synchroton light sources stems from the rate of data production from high-throughput and automated experiments in concert with traditionally slow, \emph{post hoc} analysis techniques.
In 2021 alone, it is estimated that the National Synchrotron Lightsource II (NSLS-II) will create 42 petabytes of data, with all US Department of Energy light sources data in the exabyte (1 billion gigabytes) range over the next decade.\cite{Schwarz_2020}
The developments of new tools are underscored by the increasing transition to partially of fully autonomous operation for safe and effective experiments.
Even without the possibility for autonomous operation, the beamline use remains a supply-limited resource for researchers. As such, there has been a surge on interest in optimal use of experimental resources,\cite{Zhou_2017, Roch_2018a} especially with beamline science.\cite{Maffettone_2021} 
As key emerging technologies, AI and ML enable experiments at the light source to be performed more efficiently, intelligently, and safely.\cite{Campbell_2020}
Unfortunately, the need for these tools comes with a mismatch of expertise: the predominant users of beamlines are experts in the materials of interest or analytical techniques, and not necessarily AI or computer science.

Another substantive barrier to utility is  real-time integration of AI with experimental workflows. Even with accessible and interpretable AI, the high volume data acquisition from automated and remote experiments is creating a necessity for on-the-fly monitoring and model predictions.
As experiments are commonly programmed to run ahead of time, measurements will continue indefinitely or on a fixed schedule, unless interrupted by the experimenter.
These naively automated experiments suffer from several common pitfalls, including: allocating excessive measurement time to uninteresting samples, neglecting pivotal changes in the experiment and continuing measurements during operational failures.
Real-time monitoring would solve these challenges by enabling researchers (or algorithms) to re-allocate measurement time to promising samples or parameters on-the-fly, as well as to stop or revisit an experiment that is not producing a fruitful measurement.
The initial steps toward this monitoring have been implemented as data reduction techniques that operate on raw data streams using analytical or empirical computation. These techniques  take a high dimensional data stream (2-d, 3-d, time series, etc.) and reduce it to a lower dimensional and interpretable signal.\cite{Xpdan, Abeykoon_2016}
Given the surge of new techniques and accessible software frameworks for developing AI models,\cite{scikit-learn, tensorflow, pytorch} more general interfaces are necessary to enable the growing suite of AI tools to be accessible at beamlines.  
\par

While these contemporary methods are fit to solve the immense data challenges presented during routine beamline scientific operations, there continues to exist conceptual and technical barriers the hinder beamline users and staff from readily integrating these methods.
It is worth defining what exactly is the relationship between AI, ML, deep learning, and other learning methods.
Artificial intelligence is an overarching term for any technique used to have machines imitate or approximate human intelligence and behavior.\cite{Mehta_2019} A subset of these techniques falls under the definition of machine learning, which is essentially applied statistics.
To this end, a simple and common form of ML is linear regression: fitting a line to a set of points, to subsequently use that line as a prediction for new points. The line of best fit serves as a ML model for the function of those points, to be validated when applied to new data. A particularly strong model will be predictive for new data beyond the domain of the initial training data (\emph{i.e} extrapolative).
More complex statistical models exist, and these make up the toolkit of ML. A subset of these models are considered `deep' models, which are capable of learning new mappings between ordinate spaces directly from the data, and are hailed as universal function approximators.\cite{Mehta_2019}
Critical to the scientist are the kinds of data the model consumes, and the nature and uncertainty of outputs, inference, visualizations, or directives it produces. 
\par

The technical barriers to entry occur with model building and model deployment.
The former challenge is addressed with the many accessible resources and platforms for designing AI solutions, of which we favor the Python ecosystem.\cite{Bishop_2006, scikit-learn, Mehta_2019}
It is a common case that a domain-specific AI model is developed prior to an experiment or through collaboration with technical experts external to an experiment.
In this circumstance, facile integration at the beamline is necessary for utilizing the model during the experiment.
Through three distinct relevant challenges, we will explore the different paradigms of what can be learned, with a limited focus on model details, and various operating modes of deployment.
Our principal objective is to enable the reader to understand when and how to consider AI---or alternatively when it serves limited purpose for their experiment---and demonstrate recent technological innovations that facilitate the use of AI at modern beamlines. 
\par

The challenge that remains is ensuring researchers applying these methods are accessing the best tool for their job.
Thematically, we will focus on three domains of ML:
(i) unsupervised learning as a mechanism for analyzing and visualizing unlabeled data; 
(ii) anomaly detection for identifying rare events or points in a data stream;
(iii) and supervised learning for predicting a functional labels or values associated with a data stream. 
Unsupervised learning algorithms identify and react to commonalities in  data without knowledge of the class, category, label for each datum. These approaches have been effective in reducing the dimensionality of a large dataset, and segregating physical response functions such as diffraction data \cite{Stanev_2018} and other spectra.\cite{bonnier_2012, wasserman_1997}
Anomaly detection, or outlier detection, is a reframing of unsupervised learning for identifying rare events that differ significantly from the majority of the data. The detected outliers can be scientifically intriguing as in the case of gravitational waves,\cite{Abbott_2016} or experimentally detrimental, as in the case of system failure.\cite{Borghesi_2019}
Supervised learning predicts output labels that can be discrete (\emph{classification}), such as identifying phases of matter in a dataset,\cite{Carbone_2020} or a continuous response signal (\emph{regression}) like temperature or energy.\cite{Batra_2020}
\par

Herein, we demonstrate the utility of diverse machine learning  methods for real-time monitoring of streaming data from three distinct experimental challenges at the National Synchrotron Light Source II (NSLS-II) at Brookhaven National Laboratory (BNL).
We describe these use cases pedagogically, so that they may be instructive to operators and users of the facility, opening with general instructions to overcome conceptual hurdles with developing an AI solution.
First, we demonstrate on-the-fly data segregation during a total scattering measurement, splitting a 1-d dataset into relevant components using unsupervised dimensionality reduction. This is a common challenge when conducting a diffraction experiment across phase transitions (\emph{e.g.} over a composition or temperature gradient), and allows a researcher to focus on regions near the transition.
Then, we explore the challenge of flagging a measurement when something is different from the norm established based on historical data. This unusual behavior can be caused by experimental artifacts during data collection, e.g. change of beam brightness, beam-induced sample damage, or by novel observations, e.g. a phase transition or a resonant excitation. This is particularly relevant in measurements with very sparse data points \cite{Abbott_2016} or for quality control.\cite{chou_2014}
Lastly, we solve the operational challenge of identifying failed measurements as they occur using supervised learning. In the case of X-ray absorption spectroscopy (XAS), there is a well defined feature in a measurement that when not present, indicates a failed measurement. By labeling a small set of experimental data, a supervised classification approach is shown to correctly classify new measurements. 
We close with a discussion on how each of these approaches is technically implemented and comment on the infrastructure of the Bluesky project\cite{Bluesky} for enabling everyday use of AI at central research facilities.

\section{Pipeline for developing an AI solution}
While we employed a variety of different modelling techniques and data sources in the following, the general approach to developing an AI-based solution is similar throughout.
The first step is defining the problem. By understanding which of the archetypal domains the problem falls into and the key performance metrics, one defines the approach to take with the data as well as the suite of models available to explore. Secondly, the process of data ingestion needs to be well defined. Bluesky's data model and Databroker are an example \cite{Bluesky} of a community supported framework that are suitable for solving the engineering challenge \cite{Campbell_2020} of interaction with data for AI-enhanced experiments.
\par
\begin{figure}
    \centering
    \includegraphics[width=\linewidth]{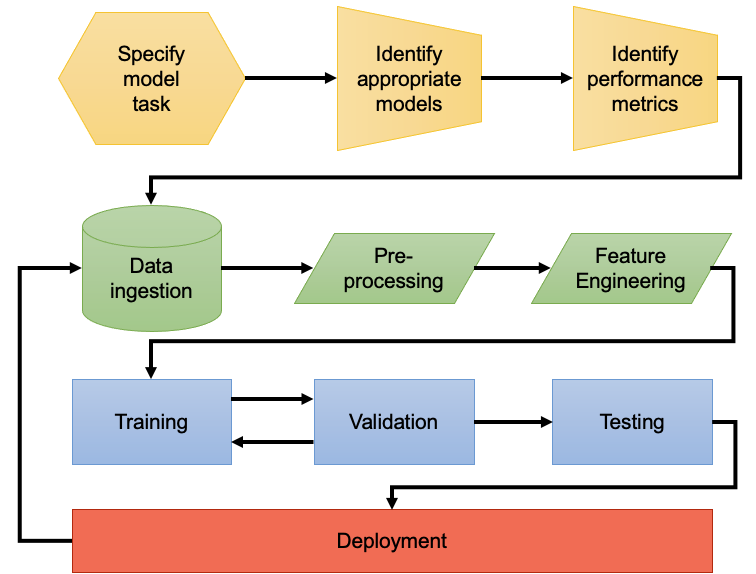}
    \caption{Flowchart describing the pipeline for developing an AI solution for a beamline science problem.}
    \label{fig:flowchart}
\end{figure}
Once the historical or active data stream is accessed, it can be prepared as input for the AI algorithms. In some cases, this requires only reformatting or rescaling the data, tailoring it for specific algorithms, where in others this amounts to performing the traditional domain-specific data reduction currently deployed at a beamline. However, it can be valuable to perform an additional data preparation step of \emph{feature engineering}: a procedure of generating a new set of calculated variables (or \emph{features}) from the original data.
New features aim to simplify the functional form of a suitable model (power transformations of a variable to fit a linear model in case of a polynomial dependence), reduce the variable range (`day'/`night' instead of a timestamp), or extract meaningful information from raw data (frequency, phase and amplitude from a wave signal of an arbitrary duration). Properly designed features can significantly improve the accuracy of the model and reduce the need for computational resources. We demonstrate feature engineering in the examples of anomaly detection (Sec.~\ref{sec:anomaly}) and supervised learning (Sec.~\ref{sec:supervised}) by using learning from a set of summary statistics of the data instead of the reduced or raw data.
\par
The processed data can then be split into training, validation, and/or test subsets to allow for effective model selection. The training dataset is used to condition (or fit) the models that are being considered. This dataset is used to minimize the models' loss functions by adjusting their respective parameters. The validation dataset is not used to train, but it is the basis to provide an unbiased evaluation of a given model's performance when comparing it to other models. The validity of each trained model is evaluated by comparing the predicted response function of the validation dataset (i.e., "model output") against the true response, which is established by scientists curating the data. A series of metrics suitable to the problem are used to quantify the performance of a model on the training and validation datesets. 
\par
Caution here is taken around the bias--variance trade-off. High bias occurs during underfitting, i.e., when the validation metrics equal or outperform the training metrics due to a model or feature set that lack the ability to express the complexity in the data. High variace occurs during overfitting, i.e., when the validation metrics significantly underperform the training metrics due to an overparameterized model that interprets noise in the training data as significant for generalization. The test dataset (or holdout set) is used to evaluate the set of final models. Care must be taken not to expose the model to the test dataset during training so as not to bias the model selection. This paradigm of splitting the data into sets is especially well suited for supervised learning tasks where labels are available. 
\par
Finally, the suite of models that are appropriate for the task is trained. Each model type has a set of adjustable hyperparameters that will impact training (the description of each is left to the resources on specific models and their implementation documentation).\cite{scikit-learn} These hyperparameters are tuned while each model is evaluated using the training and validation datasets. 
The pairings of models and hyperparameters are compared using their validation metrics. 
A few common metrics are employed in this work to evaluate models that can be expressed in terms of binary correctness:
true positives, $TP$, true negatives, $TN$, false positives, $FP$, and false negatives, $FN$.
The fraction of correct model predictions is called the accuracy,
\begin{equation}
\mathrm{Accuracy} = \frac{TP+TN}{TP+FP+TN+FN}.
\end{equation}
The precision describes the proportion of positive identifications that were actually correct,
\begin{equation}
\mathrm{Precision} = \frac{TP}{TP+FP}.
\end{equation}
In anomaly detection, precision is commonly re-framed to the false discovery rate,
\begin{equation}
  \mathrm{FDR} = \frac{FP}{FP+TP}.
\end{equation}
And the recall describes the proportion of actual positives was identified correctly,
\begin{equation}
\mathrm{Recall} = \frac{TP}{TP+FN}.
\end{equation}
Lastly, a balanced metric of the precision and recall, F$_1$ score is calculated from the harmonic mean of precision and recall,
\begin{equation}
\mathrm{F_1} = \frac{TP}{TP+\frac{1}{2}\left(FP+FN\right)}.
\end{equation}

\par
Once a suitable model has been trained, validated, and tested, it needs to be deployed. The deployment strategies vary from incorporating a fully pre-trained model into an online or offline data analysis, to fitting the model actively during an experiment using newly acquired data. Each step of the model development pipeline from problem definition to model deployment can be revisited in an iterative cycle as new data arrives or the core challenges change. 
\par

\section{Unsupervised learning}
\label{sec:unsupervised}
When a dataset has no labels to predict, or has no labels available for each datum, we turn to unsupervised learning for finding hidden patterns in the data. These methods require limited human supervision, often taking only input hyperparameters, and are commonly used for visualization of a dataset.\cite{Bishop_2006} Unsupervised methods can be categorically split between clustering and dimensionality reduction.
When confronted with unlabeled data that requires visualization or segregation, the choice of which of these approaches to use depends on the dimensionality of the data (for scaling) and framing of the problem: some algorithms will provide only groupings, while others can potentially provide meaningful information about the groups themselves. 
\par

Clustering methods are concerned with dividing data into related groups that have similar properties and/or features. These include algorithms such as expectation maximization,\cite{Dempster_1977} k-means clustering,\cite{Lloyd_1982} and hierarchical clustering.\cite{Ward_1963}
Commonly used during exploratory data analysis or to produce preliminary groupings, these methods are difficult to evaluate in their true unlabeled setting and are often ranked using a similar labeled dataset.\cite{Bishop_2006}
The choice of model is often dependent on the shape of the data distribution. Strong examples of failure modes in two dimensions is offered in the scikit-learn documentation. \cite{scikit-learn}

\par

In relation, dimensionality reduction attempts to reduce or project the data into a lower dimensional subspace that captures the core information of the data.
These include principle component analysis (PCA),\cite{Ringner_2008} singular value decomposition (SVD), \cite{Coelho_2003} non-negative matrix factorization (NMF),\cite{Geddes_2019} and deep methods such as variational autoencoders.\cite{Doersch_2016}
These methods are often used to cast a problem with many input variables down to a more manageable number of features and have found utility across the natural sciences.
One attribute underpinning their utility is the production of a series of basis vectors during the dimensionality reduction.
In the case of spectral decomposition, the non-negative basis vectors can have physical significance as end members of the dataset.\cite{Rousseeuw_1987, Geddes_2019} We use this property here in the live exploration of total scattering data via NMF, which constructs a components matrix, $\mathbf{H}$, and a weights matrix, $\mathbf{W}$, such that their product approximates the true dataset, $\mathbf{V}$, by minimizing the Frobenius norm of the difference, $||\mathbf{V} - \mathbf{WH}||_F$.

\begin{equation}
  \mathbf{V} \approx \mathbf{WH}
\end{equation}

The shape of $\mathbf{W}$ is $m \times n$, and the shape of  $\mathbf{H}$ is $n \times p$, where $m$ is the number spectra, $n$ is the length of each spectra, and $p$ is the number of components or end-members.
\par

Commonly, analytical measurements are conducted across a series of state variables, for example temperature, pressure, or composition. The combined hardware and software innovations at central facilities enable \emph{ex situ} and \emph{in situ} characterization \cite{Campbell_2020} with predetermined measurement plans. In these circumstances, large amounts of data are collected across distinct phases or other state regions of interest, with no prior knowledge of labels or transitions. It is often not until after the experiment is complete that the researcher has the opportunity to separate these regions, at which point they may be unable to explore interesting regions in more depth. We demonstrate this challenge using total scattering studies from the PDF beamline at NSLS-II, of the molten NaCl:CrCl3 (molar ratio 78:22), wherein the coordination changes of particular ions across phases impact corrosion characteristics.\cite{Li_2020} Knowledge of these materials and their corrosion characteristics is essential for their utility in molten-salt nuclear reactors.  
\par
During a temperature scan in a single sample, various crystalline and amorphous phases and their mixtures will emerge. An unsupervised method is required that can separate sets of patterns (\emph{i.e.} regions of temperature) that are distinct, thus turning a vast dataset into actionable knowledge. Various unsupervised methods can be used to segregate diffraction data using different metrics.\cite{Iwasaki_2017} Recent developments in NMF show promise for spectral functions that are positive linear combinations within mixtures.\cite{Stanev_2018, Maffettone_2021a}
NMF reduces the dataset such that each data point is described by a strictly additive mixture of relatively few non-negative end members (e.g., unique phases or components). The number of end members can be decided automatically based on other algorithmic approaches;\cite{Rousseeuw_1987} however, given the ease of calculation and knowledge of the researcher about the materials system and potentially relevant phases, it is more effective to grant the user control over this number.
Furthermore, the researcher can also focus the decomposition algorithm on a spectral range of interest. This enables a researcher to conduct rapid analysis during a variable scan and makes effective use of remaining measurement time for scientific output.  
\par
\begin{figure}
    \centering
    \includegraphics[width=\linewidth]{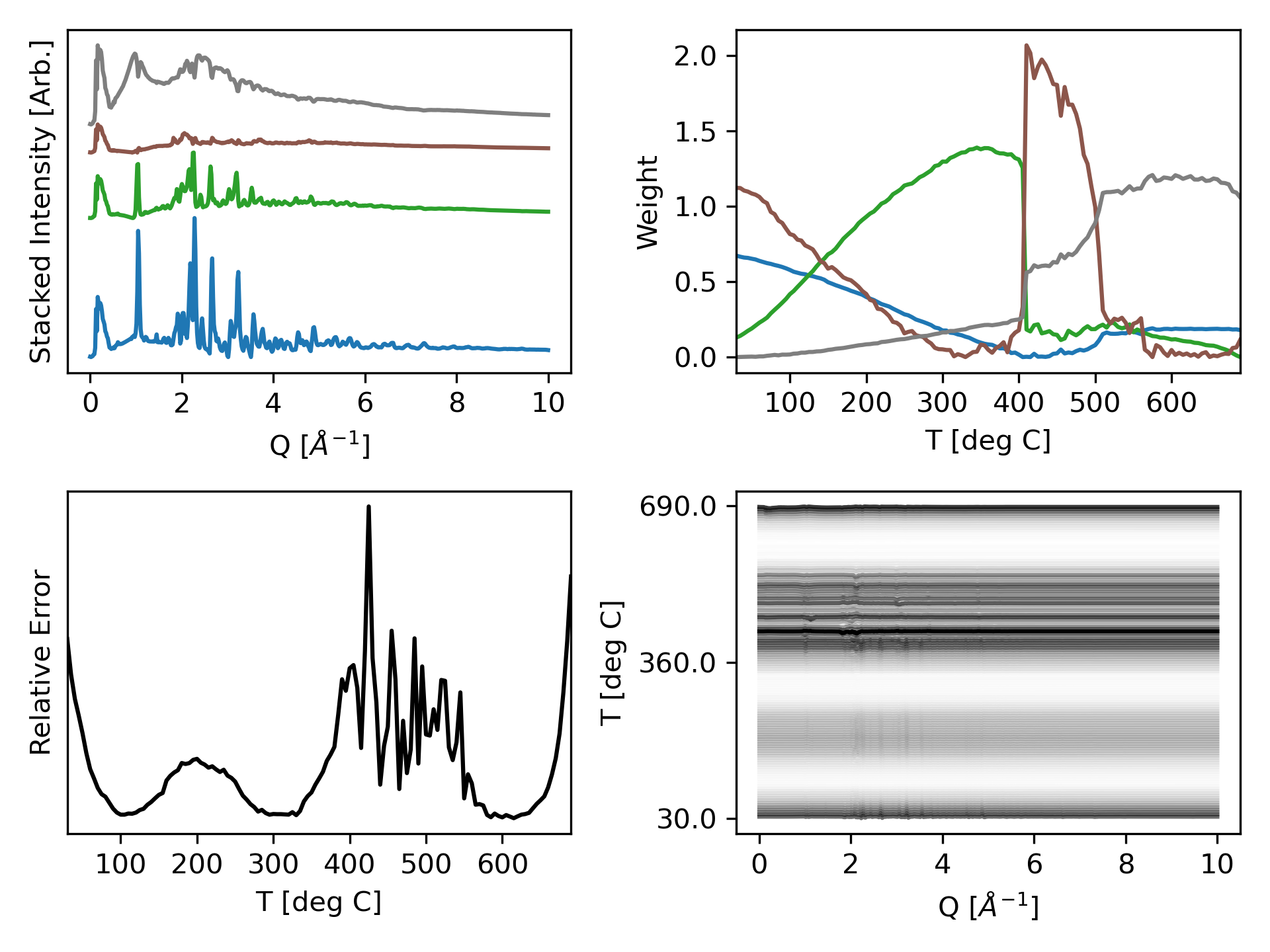}
    \caption{NMF segregates a series of spectra into a set of non-negative components, wherein the user can choose how many components are expected. (a) The resultant components used in the reconstruction of the full profile.  
    (b) The resultant components used in the reconstruction of the full profile shown with respect to the measurement temperature.
    (c) The relative error of the reconstruction with respect to each pattern at a given temperature. This shows the datum which a model does a poor job of describing the dataset.
    (d) The residual difference between the ground truth and reconstruction of each pattern with an opacity given by the reconstruction error of (c) shows where in the spectra the model is failing. 
    }
    \label{fig:PDF}
\end{figure}

We deployed NMF using Bluesky framework \cite{bluesky-website} in the study of molten NaCl:CrCl3 across a temperature range of 27--690\,$^{\circ}$C. We used the scikit-learn  implementation \cite{scikit-learn} of NMF to calculate the decomposition each time a new measurement is completed and combined the computation with a dynamic plotting using matplotlib.\cite{hunter2007matplotlib}
The implementation allows for on-the-fly monitoring and analysis of an experiment, whereas previous approaches --- even those depending on ML --- are focused solely on \emph{post hoc} analysis. The resulting display that our implementation produces is compared against a stacked plot of all of the data colored by temperature in Figure \ref{fig:PDF}.

\par
In this instance, a maximum of four end-members, and thus phases of interest, are included. The weights of each component, $W$, are shown across the temperature range, showing a smoothly varying mixture of three plausible phases in the low temperature regime, and an abrupt transition around 400 \,$^{\circ}$C. These correspond to solid mixtures and a second order phase transition to the liquid region.  Also shown is the presence of the third (green) end member in liquid regime, suggesting kinetically stabilized crystallites during melting.
Since NMF is only considering linear combinations of components, any substantial peak shifting (from changing lattice parameters or coordination) will appear as a distinct component. While some innovative models have attempted to handle peak shifting,\cite{Stanev_2018} they were not considered in this study as we constrained our focus to integrating NMF into an on-the-fly data acquisition process.
As opposed to declaring distinct phases---a task more suited for full pattern refinement \cite{TOPAS}---NMF highlights unique regions of interest in the temperature scan for interpretation. This summary is refreshed in real time with each measurement, granting the user immediate insight.  

Unsupervised ML methods are impactful beyond X-ray total scattering in the present use case. As presented, NMF enables a researcher to make effective use of precious beam time, by identifying potential experimental regions of interest and conditions for further measurement on the fly: it would be trivial to locate any second order phase transitions and conduct a subsequent scan slowly in that temperature regime.
Since advanced detectors now measure in the MHz range, data sorting is not manually feasible, and unsupervised approaches could also be used here as well.
These concepts are directly amendable to increasing automation and adaptive learning \cite{Langner_2020, Batra_2020} efforts in materials research. As implemented decomposition and clustering algorithms can be readily deployed onto other beamlines or types of measurement that produce relatively low dimensional data (1-d or small 2-d). For higher dimensional data, deep learning algorithms could be used in tandem to reduce the data or identify latent features.\cite{Doersch_2016}

\section{Anomaly detection}
\label{sec:anomaly}
Anomaly detection algorithms aim to identify unexpected observations that are significantly different from the remaining majority of observations. 
Such algorithms are used for many different tasks, including credit card fraud detection,\cite{Tran_2018} discovering unusual power consumption,\cite{chou_2014} and identifying cyber security threats.\cite{Bhuyan_2013}
Isolating anomalous instances can be accomplished using supervised or unsupervised learning. Further detailed in Section~\ref{sec:supervised}, supervised algorithms require the training data to be labeled and have a proper distribution of the different types of abnormal cases that can be encountered. However, the knowledge of potential types of anomalous data is often not available before the data are taken. Unsupervised algorithms, on the other hand, do not assume the knowledge of types of possible irregularities. They are based on the presumption that majority of the data is normal with anomalies being rare and divergent from the ordinary data. Such algorithms tend to learn the distribution of the normal data according to specific hyperparameters. Sample points that are unlikely to come from this distribution are labeled as outliers. In the circumstance when all data outside the normal expected signal cannot be predicted and labeled, but still need to be identified, unsupervised anomaly detection is an incredibly useful tool.
\par

\begin{figure}
   \centering
    \includegraphics{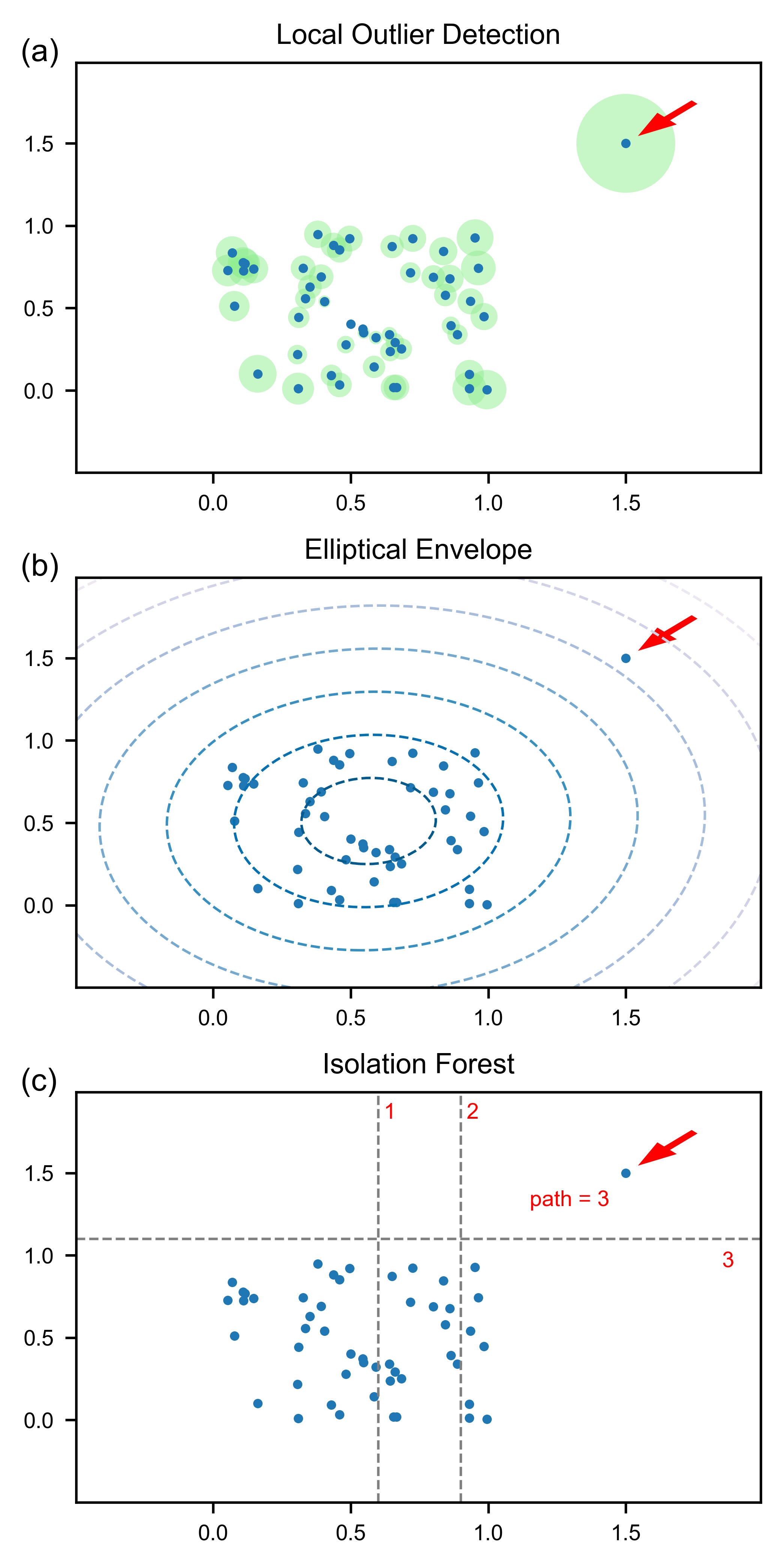}
    \caption{Graphical description of anomaly detection algorithms: (a) local outlier detection, (b) elliptical envelope, and (c) isolation forest.}
    \label{fig:anomaly_algo}
\end{figure}

Here, we focus on three unsupervised algorithms: local outlier detection \cite{Breunig2000} (LOD), elliptical envelope \cite{Rousseeuw1999} (EE) and isolation forest \cite{Liu2008} (IFT).
The LOD algorithm identifies the regions with similar local density of points based on several nearest neighbors (Fig.~\ref{fig:anomaly_algo}a). The points with local density smaller than their neighbors are identified as outliers. The degree of certainty with which a point is attributed to outliers depends on the number of nearest neighbors considered---an additional hyperparameter of the model.
The EE algorithm assumes that the normal data are centered around a single point and fits a multidimensional ellipsoid around the center (Fig.~\ref{fig:anomaly_algo}b). Whether or not each point is considered an outlier is based on the Mahalanobis distance between the point and this elliptical distribution. Exploratory analysis of the principle components shows that the normal data in our case constitute a single cluster, though its shape is not close to elliptical in some planes.
The IFT algorithm has an isolation tree as its basic structure. Such tree is built by randomly selecting a variable and a split point until each leaf of the tree contains only samples with the same values. A path to the leaf is equal to the number of partitioning necessary to isolate the sample. The length of a path to a point, averaged over the collection of the trees (the forest), is the metric used to determine if it is an outlier (Fig.~\ref{fig:anomaly_algo}c). The algorithm is known to outperform other methods for variety of cases \cite{}, though is can be computationally expensive for high-dimensional data.

Synchrotron user facilities can benefit from integrating anomaly detection tools into their operations. Often in a measurement setting, it is necessary to highlight when a result is different from what is expected. In some cases, this amounts to recognizing an equipment failure early, thus allowing the researcher to react promptly. In other instances, this would take the form of finding new or interesting data points within a larger dataset. Both tasks would normally require constant monitoring of collected signals by the researcher. Anomaly detection algorithms can be integrated in the online data analysis for prompt evaluating of the measurements, reducing the need for human efforts.   

\begin{figure*}
    \centering
    \includegraphics{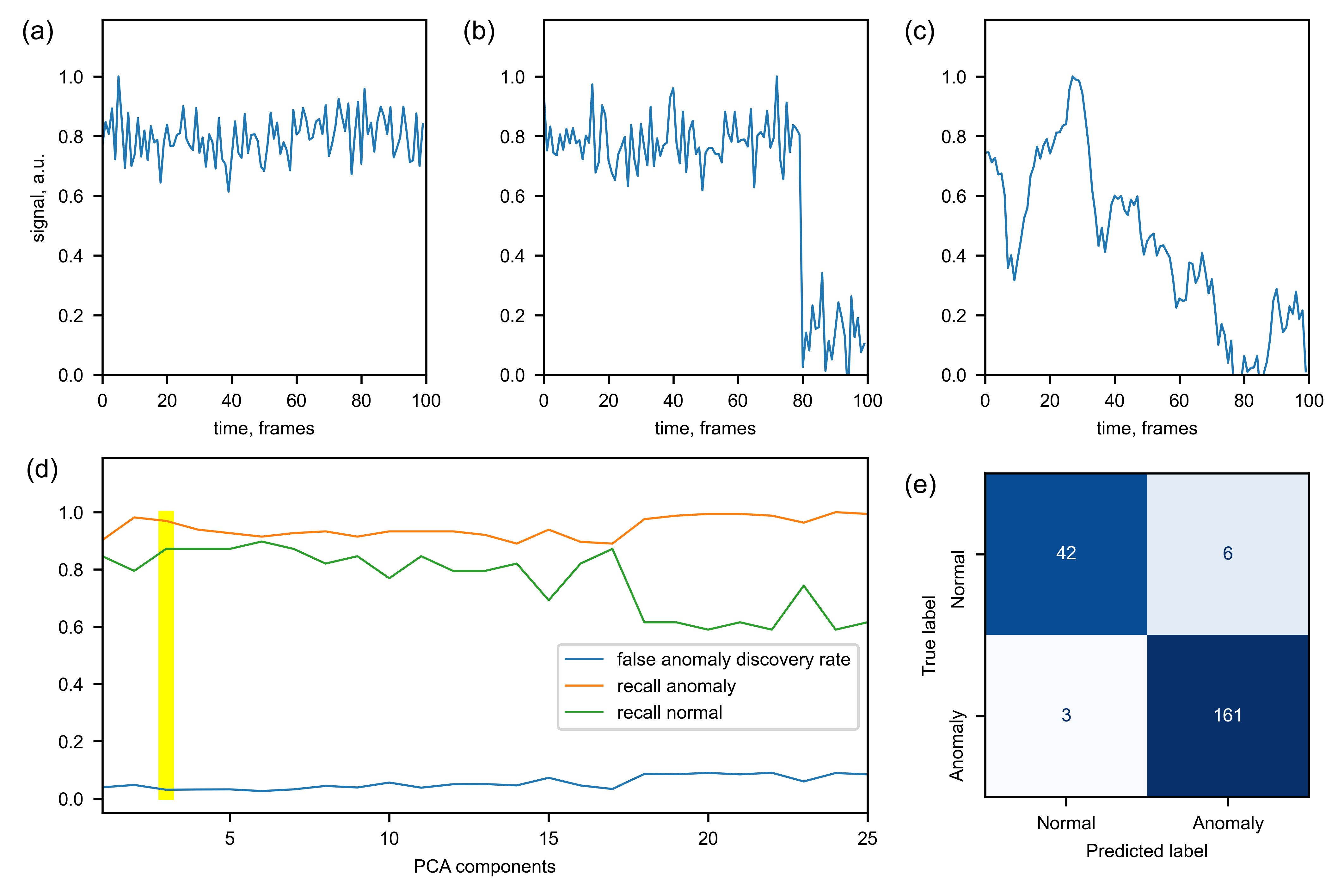}
    \caption{(a) Normal series. (b) Discontinuity in the data. (c) Strong fluctuations in the data. (d) Performance of the EE model on validation data set for different number of principle components. The region with the best model performance is highlighted in yellow. (e) Confusion matrix that reflects the performance of the final EE model on the test set. }
    \label{fig:Anomaly}
\end{figure*}

We built an anomaly detection toolkit for the time series collected for the X-ray Photon Coherent Scattering (XPCS) experiments.\cite{Shpyrko_2014, Sinha_2014} During the measurements, series of scattering images (frames) are recorded by the 2D area detector (e.g. CCD). As part of the analysis, the photon intensity for a group of pixels is autocorrelated over time since the decorrelation of the speckles' intensity is reflective of the inherent sample's dynamics (Fig.~\ref{fig:Anomaly}a-c). Consequently, the events like a sample motion, the X-ray beam drift, or changes in the beam intensity can lead to artifacts in the correlation functions.\cite{Campbell_2020} In some occasions, changes of the scattering peak’s position or intensity can be due to the intrinsic properties of the samples.
As the anomalous events encountered during the data collection, they should be investigated by a researcher, who can dynamically adjust the experimental plan or conditions. Since experiments last extended periods of time (on the order of days) and are controlled by pre-assembled plans, it is critical to have an automated tool that alarms the researcher about anomalous observations, so they may target the most critical experimental conditions first and then make appropriate decisions regarding subsequent measurements and analysis.
\par 

The data for this work were collected by processing results of previous measurements at the CSX beamline at NSLS-II. For multiple regions of interest at the 2D detector for each frame we calculate 6 time series: the total intensity, the standard deviation of binned pixels' intensity counts, the center of mass coordinates and its standard deviations for both directions. These variables are chosen because they can be directly calculated and reported by the detector's control software during an experiment and circumvent the need for more time consuming post-processing of scattering images. While such algorithms do not require the labeled data for the training, we annotate a dataset for the purpose of evaluating the models’ performance. Each example is labeled as ‘normal’ or ‘anomalous’ based on the expert knowledge. Figure \ref{fig:Anomaly}  illustrates how an intensity time series can look in a normal (a) and anomalous (b,c) cases. In a normal case, all considered experimental parameters are (almost) stationary, while anomalous cases may contain sudden jumps or significant drifts of the parameters’ values. The duration of each time series ranges from tens to thousands of frames. To be processed by the anomaly detection algorithms, the series need to be converted to a set of variables of a fixed length. 

\par 
We engineered a set of 93 features to capture the statistical diversity of the variable length data. The derived features include (i) standard deviation to mean ratio within the series, (ii) autocorrelation coefficients up to 4th order, (iii) ratio of a parameter’s standard deviation to the standard deviation of its first time derivative, (iv) difference of parameter values at the beginning and the end of the scan. The feature engineering aims to highlight the lack (presence) of trends and discontinuities in normal (anomalous) cases and can be helpful in other ML tasks involving sequential data. The features were calculated from the times series following two preprocessing steps: centering the series around their mean values and normalizing by the mean values. The second step was not included for the series related to the intensity peak positions as their absolute displacement values can be indicative of outliers and thus should be preserved.  

As EE and LOD use Euclidean distance measure in the base of their algorithms, it is likely that the models do not perform well in a high-dimensional space. Moreover, the number of variables is comparable to the number of examples in our train set, increasing the potential risk of over-fitting.  To address these concerns we control the dimensionality of the data. We employ an unsupervised dimensionality reduction technique, principle component analysis (PCA),\cite{Ringner_2008} similar to that presented in Section~\ref{sec:unsupervised} to reduce the size of our engineered feature vectors. PCA is an orthogonal linear transformation that transforms the data to a new coordinate system such that the greatest variance by some scalar projection of the data comes to lie on the new coordinates. These scalar projections are used as our reduced dimensions. 
\par 
The data are divided into exclusive sets for training, validation and testing. Only the `normal' data are used for training the models and the training set contains 80\% of all normal examples – the rest is evenly divided between the validation and the test set. In doing so, we ensure that the assumption of the models about majority of the data being 'normal' is satisfied. Since the considered models do not rely on the data labels for training process, a significant presence of anomalous examples in the training set could deteriorate a model performance.
The validation dataset is used for identifying optimal hyperparameters of the models. In addition to the 10\% of the normal examples, it contains 50\% of the anomalous examples. The rest of the data belong to the test set, which is used for the final models’ assessment. The model performance can be evaluated through various parameters calculated from the confusion matrix (Fig.~\ref{fig:Anomaly}). Recall \emph{R} reflects the rate of correctly identified normal (anomalous) labels among all normal (anomalous) examples and false anomaly discovery rate \emph{FDR} reflects the ratio of incorrectly identified labels among all examples labeled as anomaly.   
\par 
The key hyperparameters we tune across all models are the dimensionality of the input signals (the number of principle components) and the contamination level. The contamination level is the percentage of anomalous examples in the train set. Despite only normal data being in the training set, we let a small portion of them to be identified as anomalous, i.e. having false anomalous labels, in expectation that an actual outlier will be even further away from the main cloud of normal data and thus correctly identified by the model. This approach prioritizes having false positives (alarms being raised prematurely) over false negatives (alarms which should have been raised being missed). The hyperparameters of the models are optimized by maximizing the products \emph{R*FDR} for the normal and anomalous data in the validation set. An example of selection of the optimal number of the principle components is shown in \ref{fig:Anomaly}(d). 
\par  
The results of our performance comparison across three anomaly detection models are shown in Table~\ref{tab:Anomaly}. LOD has the least percentage of incorrectly labeled normal data, but it slightly under-performs in identifying anomalies comparing to other algorithms. Comparing to the LOD, the IFT algorithm demonstrates better results in correctly identifying the anomalous test data, but it mislabels more of the normal examples than other algorithms do. In our case, the EE algorithm has the best performance when considering both recall and false discovery rate. Depending on the application priorities, the threshold value of the model's metric can be adjusted to reduce either false positive or false negative outcomes.

\begin{table}
  \caption{Results of the unsupervised algorithms on the test set.}
  \centering
    \label{tab:Anomaly}
    \begin{tabular}{c|c|c|c}
         Models & LOD & EE & IFT\\\hline
         Recall Anomaly & 0.92  & 0.98 & 0.98 \\
         False Anomaly Discovery Rate   & 0.026  & 0.036 & 0.042 \\
    \end{tabular}
\end{table}

We demonstrate that anomaly detection algorithms can be an effective tool for identifying unusual time series in XPCS measurements. Automatic flagging of such observations helps optimize the workload of XPCS researchers, freeing them from the necessity of manually evaluating every dataset.
The innovations here including feature engineering, dimensionality reduction, and online unsupervised anomaly detection are not limited to applications in XPCS or even to time series. The sequence of these methods could be applied directly to any one-dimensional equally spaced data arrays, where the order of observations is important, e.g., spectra, line cuts of two-dimensional images, temperature series. More generally, the model is applicable for filtering out artefacts in a set of repetitive measurements performed \cite{cookson_2006,gati_2014} to obtain appropriate statistics in case of a weak signal, which can include higher order dimensions when suitable feature engineering is employed. 

\section{Supervised classification}
\label{sec:supervised}
Supervised learning is a very common task in science and engineering and based on the same principles as standard fitting procedures or regression. Its core objective is to find an unknown function that maps input data to output labels. When those labels are discrete, this is considered a classification task, and when those labels are continuous, it is considered a regression task.
A particularly desirable outcome of supervised learning is transferability: a model trained on a one dataset should be predictive on another, and not simply interpolative. As such, it is advantageous to perform feature engineering---which biases the generalization of the approach to the engineer's discretion---or utilize deep approaches that 'learn' the proper featurization.
In general, a problem can be cast as a supervised learning problem if there is labeled data available, and that data format can be mapped as an input for the available algorithms. 
\par
The broad impact of supervised learning is undeniable, impacting technologies in our daily lives through image classification,\cite{Lu_2007} speech recognition,\cite{REDDY_1990} and web-searches. \cite{Pazzani_1997}
However, in the domains of applied materials science and crystallography, these contemporary approaches have accelerated physical simulations,\cite{Mehta_2019} property prediction,\cite{Batra_2020} and analytical techniques such as diffraction \cite{Lee_2020} and microscopy.\cite{Kaufmann_2020}
Each of these advances is underpinned by a variety of models, some of which require deep learning to accomplish. Model selection for supervised learning is dependent on both the size of the labeled dataset, and the dimensionality or shape of the data.\cite{Batra_2020} In general with small datasets ($<$10,000 points), it is advisable to consider statistical ML algorithms over deep learning for transferable predictive performance that does not over fit. 

\par
In many circumstances there is a stark and identifiable contrast between `good’ and `bad’ data during a materials analysis measurement. Where we use the term `good’ to describe data that is  ready for immediate interpretation, and `bad’ to describe data that may be uninterpretable or which merits human intervention prior to being ready for interpretation. Bad data stems from a variety of sources, including but not limited to weak signal-to-noise ratio, improper sample alignment, or instrumentation failure. Contrary to the experimental situations we presented for anomaly detection (Sec.~\ref{sec:anomaly}), these bad data are well defined and can readily be labeled. However, there is a pressing need for on-the-fly analysis to identify when 'bad' data arises during an automated experiment, so as to enable rapid intervention. In this case, the supervised learning approaches that were not well suited for anomaly detection are useful.

\par
\begin{figure}
    \centering
    \includegraphics{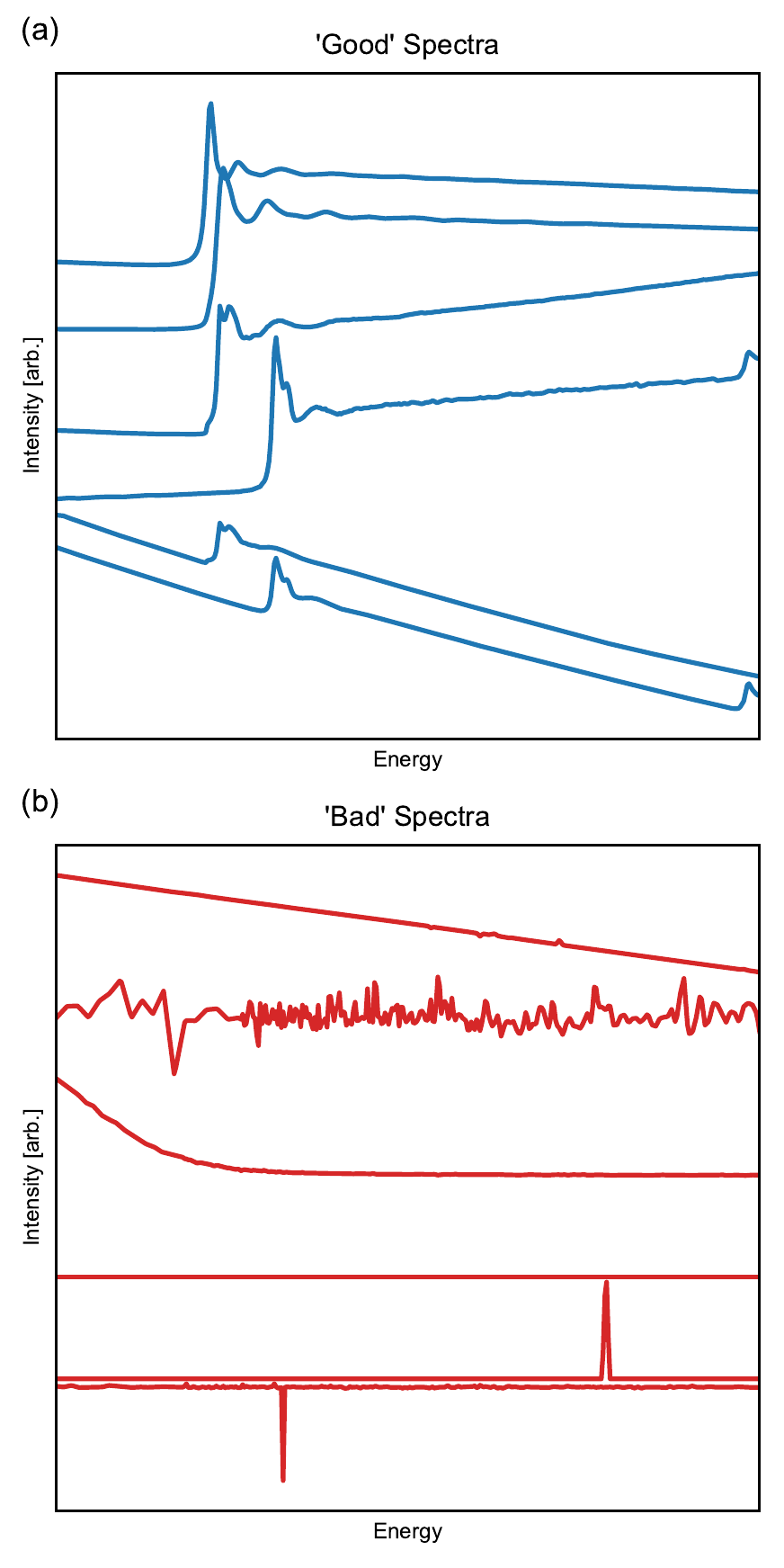}
    \caption{(a) XAFS measurements considered good, or ready for interpretation by an expert, will contain a rising edge somewhere along the scanned energy, followed by the oscillatory fine structure. (b) Data that are considered bad, or requiring experimental intervention by an expert, will not contain an absorption edge and often present as noise.}
    \label{fig:BMM}
\end{figure}

At the Beamline for Materials Measurement (BMM) at NSLS-II, X-ray absorption fine structure (XAFS) is routinely measured \emph{via} X-ray absorption spectroscopy (XAS) in a high-throughput automated experiment. The XAFS measurement varies the incident photon energy to measure the energy-dependent, X-ray-absorption cross section, which provides a direct measurement of valence and other chemical information and which may be analyzed to recover details of local partial pair distribution functions.\cite{jeroen16:_x_ray_absor_x_ray_emiss_spect} XAFS is regularly measured in two modes at a hard X-ray beamline like BMM. In transmission, the optical opacity of the sample is measured by the attenuation of the incident beam intensity as it passes through the sample. Here, the absorption cross section changes dramatically as the energy of the incident beam is scanned through the binding energy of a deep-core electron, resulting in the emission of a photo-electron and the creation of a short-lived core-hole. In fluorescence, the absorption cross section is determined by measuring the emission of the secondary photon produced when the core-hole created by the photo-excitation of the deep-core electron is refilled by the decay of a higher-lying electron. The energy range of this scan depends on the chemistry and composition of the sample and on other experimental considerations. To make the measurements accessible to non-sequential ML algorithms, every spectra is down-sampled to contain 400 members; however, the energy bounds of the spectra are not adjusted. The results of these experiments and processing are thus a set of 1-d vectors with 400 members, with typical `good' spectra shown in Figure~\ref{fig:BMM}(a).

\par

The spectra are easily labeled as good or bad data, as the good data will undergo a sharp and significant change in intensity (called an `absorption edge' in XAFS), while the bad data (Figure~\ref{fig:BMM}(b))---regardless of root cause---will lack an absorption edge and take the appearance of random noise or featureless background signal.
Regardless of the ease of this pattern recognition task, 
the current approach for good/bad classification requires human intervention and judgement, which is not ideal for remote, high-throughput, and/or overnight data collection. A ML-based classification will make for more efficient use of the beamtime.
From a set of previously collected and labeled spectra, we applied a suite of classification models with some achieving 100\% accuracy on an unseen test-set. We considered Random Forrest (RF),\cite{breiman_2001} Support Vector Machine (SVM),\cite{cortes_1995} Multi-layer perceptron (MLP),\cite{rumelhart_1985} k-Neighbors,\cite{fix_1989} and Gaussian Process (GP) classifiers.\cite{gibbs_2000} In all cases, we used the default hyperparameters in the scikit-learn implementation, except for MLP models where we reduced the default hidden nodes to 10.
\par
 
We compared the performance of these models across different splits of the labeled dataset (Table~\ref{tab:BMM}). In the first approach, a set of 711 data from transmission and fluorescence data of variable quality was randomly split into training and validation sets (80\% training and 20\% validation). We refer to these approach as uniform validation. In the second approach (unique validation), data from a set of 'very good' measurements was retained for the validation set with 10\% of the remaining data sampled for validation (156 total). This unique validation approach allows for testing of the extensibility of models, that is, how well they will behave on data outside of the scope of training. In both approaches, we explored each model's performance on the raw normalized spectra and on a set of statistical features that were calculated from the spectra and their first derivatives: (i) autocorrelation coefficients for lag 1--4, (ii) mean of the first 5 values, (iii) mean of the last 5 values, (iv) mean of the intensity, (v) standard deviation of the intensity, (vi) sum of the intensity, and (vii) location of the intensity maximum. These features were normalized by the maximum for the training data of each feature. 
 
\begin{table}
    \caption{Binary classification results from a suite of models applied to two dataset splits. Most models using the raw spectra fail to generalize to an unseen experiment in the unique validation; however, using engineered features from the statistics of the spectra enables more robust generalization.}
    \label{tab:BMM}
    \begin{tabular}{c|c|c|c|c}
         & \multicolumn{2}{c|}{Raw Spectra} & \multicolumn{2}{c}{Engineered Features} \\ \hline
         & Uniform & Unique & Uniform & Unique\\
         &  Validation & Validation & Validation & Validation\\
         Models & F1-Score & F1-Score & F1-Score & F1-Score \\ \hline
         RF & 0.986  & 0.829 & 0.990 & 0.874 \\
         SVM & 0.995  & 0.807 & 0.990 & 0.982 \\
         MLP  & 1.00  & 1.00 & 0.986 & 0.957 \\
         k-Neighbors & 0.995  & 0.807 & 0.990 & 0.947 \\
         GP & 0.990  & 0.803 & 0.986 & 0.988 \\
    \end{tabular}
\end{table}

Based on the results in Table~\ref{tab:BMM}, the challenge of effective representation becomes apparent. In the case of using the raw spectra, only the MLP models are able to make accurately predictions on data from new experiments. However, when the ML algorithms are fed derived features that capture the most important information, the models can be more effectively generalized to new data. The models trained on raw spectra fail to classify spectra with rising edges of new shapes or in different positions. This lack of generalization is unsurprising because these models do not create their own abstractions, whereas when abstractions are provided by feature engineering, the models become more useful beyond the training data. Other approaches to creating abstractions without biased feature engineering exist in the field of deep learning. Convolutional neural networks trained on the raw spectra dataset, similar to the multi-layer perceptrons,  approach 100\% validation accuracy where shallow models fail. These approaches are beyond the scope of this paper, being less accessible to the average scientist; however, their success underscore the value of feature engineering with expert knowledge since those features can be `learned' by deep algorithms.

\section{Deployment Interfaces}
The final component of the AI pipeline referenced in Figure~\ref{fig:flowchart} is deployment. 
The complexity of steps involved into a model deployment can vary considerably depending on the application. The simplest deployment strategy is to provide a pre-trained model to a user. While this approach has its advantages, such as ease of testing and flexibility of workflow modification, they also require a lot of user intervention to utilize the model via managing file transfer, data inputs and outputs, as well as the interpretation of the model output.   
However, the most user friendly interfaces imply that the complexity of the deployment is unlimited because the AI tools are seamlessly integrated into existing workflows that can make for superior experiments  because the AI is no additional "work" for the user. Such an interface enables both
human-in-the-loop operation,\cite{Stach_2021} and completely autonomous experiments.\cite{Roch_2018a, Burger_2020}
While many of the beamlines at NSLS-II use similar interfaces, it is common for particular experiments or beamlines to have bespoke software solutions built on or interfacing with common frameworks, such as Bluesky\cite{bluesky-website} or Ophyd.\cite{ophyd}
Here we outline how each of the proceeding sections was implemented, to demonstrate the diverse integration modes across the facility. 
\par

Firstly, it is useful to have a generic interface to expect with AI models, so that similar models can be deployed in different experimental processes regardless of other design decisions.
Following recent work in adaptive experiments at NSLS-II,\cite{bluesky-adaptive-github, bluesky-adaptive-tutorial} we implemented all of these models with an
\textbf{tell}--\textbf{report}--\textbf{ask} interface.
That is, each model was part of some object that had a \textbf{tell} method to tell the model about new data,  a \textbf{report} method to generate a report or visualization, and an \textbf{ask} method to ask the model what to do next. 
While the latter method is required with adaptive learning in mind, it enables simple adaptations such as a model detecting an anomaly and wishing to pause the experiment. 
This generic interface suits most needs for AI at the beamline, and allows users to `plug-and-play' models they have developed without considering how the data is being streamed or other communication protocols. 
A complete tutorial using the \textbf{tell}--\textbf{ask} components to deploy multiple AI models can be found at reference \citenum{bluesky-adaptive-tutorial}, and all the models contained here are available (see Code Availability statement).
\par
The example deployment of NMF demonstrates how the \textbf{tell}--\textbf{report}--\textbf{ask} interface can be used without any data streaming.
At the PDF beamline at NSLS-II, raw diffraction images from a 2-d area detector are streamed as documents to data reduction software, that produces data for scientific interpretation as a 1-d pattern stored in a file system locally or on a distributed server. 
Here, we employ the \textbf{tell}--\textbf{report}--\textbf{ask} interface inside a file system watcher. 
The model is told about new data each time new files appear, and subsequently generates a new report, \emph{i.e.} the visualization shown in Figure~\ref{fig:PDF}. 
Due to the inexpensive nature of updating the NMF model, this gives the researcher a developing model and visualization over time. 
This example also shows how to include both model training and evaluation in-line with an experimental data stream.
\par
It is possible to train and use a model completely offline using the corpus of data generated over an extended period of time without the need to constantly update the model after a new measurement. For anomaly detection model, we used data from multiple experimental measurements, separated by an expert into normal and anomalous groups. The training of the model and selection of the best performer is done in a Jupyter Notebook environment because simple pipelines are all that is required for development and testing. The model can be deployed for both online data streaming application and for offline analysis. Its self-contained simplicity allows a user to insert the model to work best in their preferred workflow. In the examples accompanying this work, we demonstrate \textbf{tell}--\textbf{report}--\textbf{ask} interface for the file system. With such an interface, the model can access the result files from a folder and return a prediction of whether the measurement considered an anomaly.  The output of the model can be utilized by subsequent AI-guided analysis and results extraction.\cite{konstantinova_2021} The model does not have to be updated after each new measurement is added to the folder and routine model updates can be scheduled when sufficient amount of new data are acquired and labeled. The process of model update and application can easily be automated using Papermill.\cite{papermill}
\par
In our deployment of supervised learning for identifying failed measurements at BMM, we constructed a part of Bluesky plan---a callback---to publish the report from the model onto Slack, a common business communication platform, in the form of emojis.
This enables remote monitoring of an experiment for potential failures, as well as a timeline of those failures. 
We use a class with the  \textbf{tell}--\textbf{report} interface, separate from the callback designed by the beamline scientist, so that the models and report styles can be easily interchanged. 
The ML models can either be loaded from disk or retrained from a standard dataset at the start of a each experiment. Each time a measurement is taken, a report is generated based on the model classification and passed to the callback that processes the report for Slack.
This deployment shows how the same interface used for monitoring directories in a file system, can be quickly linked to streaming data, and publish results to the internet or a chat service. 

\section{Conclusions}
AI opens opportunities for making many beamline experiments more efficient in various aspects from data collection and analysis to planning next steps.
As the rate of data production continues to increase with new high-throughput technologies, and remote operations requirements grow, new analytical tools need to be developed to accommodate this increased flux of data in a distributed manner.
Herein we tackled three unique experimental challenges at NSLS-II that fall under individual archetypes of machine learning: unsupervised segregation, anomaly detection, and supervised classification. 
We integrate non-negative matrix factorization to separate key components of total scattering spectra across a temperature driven phase transition. 
Secondly, we deploy anomaly detection to warn a user of substantial changes in the time evolution of XPCS data. And lastly, we train a supervised binary classifier to separate good data that is ready for immediate analysis and bad data that requires experimental intervention during an XAFS experiment. 
Use of these AI methods is aimed to increase scientific outcome of the experiments and does not rely on large-scale computational resources or extended software development skills. Each of the models could be trained on a personal computer in a matter of minutes or even seconds. Open-source Python libraries, such as scikit-learn,\cite{scikit-learn} make encapsulated implementation of elaborate algorithms available for researchers from wide range of disciplines. It is the researcher with domain expertise in physics, chemistry or material science that tailors the models for specific applications.

Beyond the scope of this work, yet still relevant to beamline science are adaptive learning and reinforcement learning.\cite{Bruchon_2020}
Adaptive learning is an extension of supervised learning where the algorithm can ask for more data to improve its model, and has been used in experimental optimization and search.\cite{Burger_2020, Hase_2018, Roch_2018a, Noack_2019}
Reinforcement learning approaches a related task of learning an optimal policy given a reward and penalty structure. This has recently been demonstrated for optimizing beamline operations and resource allotment.\cite{Bruchon_2020, Maffettone_2021}
Deploying these techniques at a beamline are significant enough to warrant their own study,\cite{Noack_2019, Maffettone_2021} albeit the tools we develop here are designed with adaptive protocols in mind. 

The integration of each considered model into the Bluesky Suite for experimental orchestration and data management underpins their accessibility to beamline users and staff that are unfamiliar with ML, and extensibility to new applications. These extensions include similar thematic data challenges at different experiments and algorithmic development to incorporate adaptive experiments which depend on the feedback from ML.\cite{Li_2020}
Given this framework and the scientific python ecosystem, there are boundless opportunities for further applications of these and different ML approaches in high-throughput and distributed experimental feedback loops. 
\par


\section*{Code Availability}
The source code and data to reproduce the examples in this work is available at \href{github.com/bnl/pub-ML_examples/}{github.com/bnl/pub-ML\_examples/}.

\begin{acknowledgments}
This research used the PDF, CSX, and BMM beamlines of the National Synchrotron Light Source II, a U.S. Department of Energy (DOE) Office of Science User Facility operated for the DOE Office of Science by Brookhaven National Laboratory (BNL) under Contract No. DE-SC0012704 and resources of a BNL Laboratory Directed Research and Development (LDRD) projects 20-032 ''Accelerating materials discovery with total scattering via machine learning'' and 20-038  ''Machine  Learning  for  Real-Time Data Fidelity, Healing, and Analysis for Coherent X-ray Synchrotron Data''.  We would like to acknowledge Anthony DeGennaro who is a co-PI for LDRD 20-038 from BNL Computer Science Initiative (CSI).
\end{acknowledgments}


\bibliography{bibliography} 
\bibliographystyle{rsc} 

\end{document}